\documentclass[sigconf]{acmart}
\usepackage{ragged2e}
\usepackage{latexsym}
\usepackage{enumitem}
\usepackage{xspace}
\usepackage{xcolor, soul}
\usepackage{multirow}
\sethlcolor{green}
\usepackage{multicol}
\usepackage{hyperref}

\AtBeginDocument{%
}

\copyrightyear{2024}
\acmYear{2024}
\setcopyright{rightsretained}
\acmConference[KDD '24]{Proceedings of the 30th ACM SIGKDD Conference on Knowledge Discovery and Data Mining}{August 25--29, 2024}{Barcelona, Spain} 
\acmBooktitle{Proceedings of the 30th ACM SIGKDD Conference on Knowledge Discovery and Data Mining (KDD '24), August 25--29, 2024, Barcelona, Spain}
\acmDOI{10.1145/3637528.3671745}
\acmISBN{979-8-4007-0490-1/24/08}

\makeatletter
\gdef\@copyrightpermission{
  \begin{minipage}{0.3\columnwidth}
    \href{https://creativecommons.org/licenses/by-nc-sa/4.0/}{\includegraphics[width=0.9\columnwidth]{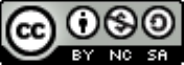}}
  \end{minipage}\hfill
  \begin{minipage}{0.7\columnwidth}
   \href{https://creativecommons.org/licenses/by-nc-sa/4.0/}{work is licensed under a Creative Commons Attribution-NonCommercial-ShareAlike International 4.0 License.}
  \end{minipage}
  \vspace{5pt}
}
\makeatother
\begin{document}

\title[\our: Ontology-Guided and PLM-Assisted Fine-Grained Entity Typing]{\our: Ontology-Guided and Pre-Trained Language Model Assisted Fine-Grained Entity Typing}


\author{Tanay Komarlu}
\affiliation{%
  \institution{University of Illinois Urbana-Champaign}
  \city{Urbana}
  \state{Illinois}
  \country{USA}
}
\email{komarlu2@illinois.edu}
\author{Minhao Jiang}
\affiliation{%
  \institution{University of Illinois Urbana-Champaign}
  \city{Urbana}
  \state{IL}
  \country{USA}
}
\email{minhaoj2@illinois.edu}
\author{Xuan Wang}
\affiliation{%
  \institution{Virginia Tech}
  \city{Blacksburg}
  \state{Virginia}
  \country{USA}
}
\email{xuanw@vt.edu}
\author{Jiawei Han}
\affiliation{%
  \institution{University of Illinois Urbana-Champaign}
  \city{Urbana}
  \state{Illinois}
  \country{USA}
}
\email{hanj@illinois.edu}

\ccsdesc[500]{Computing methodologies~Information extraction}
\ccsdesc[500]{Computing methodologies~Natural language processing}
\ccsdesc[300]{Computing methodologies~Language resources}

\keywords{Fine-Grained Entity Typing, Zero-Shot Entity Typing, Masked Language Model Prompting, Natural Language Understanding}

\newcommand\myeq[1]{\stackrel{\textnormal{#1}}{=}}
\newcommand{\correct}[1]{{\color{blue}#1}}

\newcommand{\our}{\textsc{OntoType}\xspace}
\begin{abstract}
Fine-grained entity typing (FET), which assigns entities in text with context-sensitive, fine-grained semantic types, is a basic but important task for knowledge extraction from unstructured text. 
FET has been studied extensively in natural language processing and typically relies on human-annotated corpora for training, which is costly and difficult to scale.
Recent studies explore the utilization of pre-trained language models (PLMs) as a knowledge base to generate rich and context-aware weak supervision for FET. 
However, a PLM still requires direction and guidance to serve as a knowledge base as they often generate a mixture of rough and fine-grained types, or tokens unsuitable for typing.
In this study, we vision that an ontology provides a semantics-rich, hierarchical structure, which will help select the best results generated by multiple PLM models and head words. 
Specifically, we propose a novel \emph{annotation-free, ontology-guided} FET method, \our, which follows a type ontological structure, from coarse to fine, ensembles multiple PLM prompting results to generate a set of type candidates, and refines its type resolution, under the local context with a natural language inference model.
Our experiments on the Ontonotes, FIGER, and NYT datasets using their associated ontological structures demonstrate that our method outperforms the state-of-the-art zero-shot fine-grained entity typing 
methods as well as a typical LLM method, ChatGPT. 
Our error analysis shows that refinement of the existing ontology structures will further improve fine-grained entity typing.
\end{abstract}

\maketitle

\section{Introduction}

Fine-grained entity typing (FET) is a basic but important task for knowledge extraction and text content understanding/analysis. 
Assigning fine-grained semantic types to parsed entity mention spans based on the local context enables effective and structured analysis of unstructured text data, such as entity linking \cite{Weld_Ling2012, onoe2020interpretable}, relation extraction \cite{koch2014type}, and coreference resolution \cite{onoe2020interpretable}.

\vspace{0.2ex}
\noindent
\textbf{Example 1.} Given a sentence $S_1$: ``\emph{Sammy Sosa got a standing ovation at Wrigley Field.}'' and a parsed entity mention span ``\emph{Sammy Sosa}'' in the sentence, an FET method aims to assign it not only the coarse-grained type ``Person'' but also the fine-grained types ``Athlete'' or ``Player''.
\vspace{0.2ex}

FET on large text corpora is a challenging task due to (1) the high cost of obtaining a large amount of human-annotated training data, especially in dynamic and emerging domains, and (2) inaccurate annotations due to (i) different annotators marking concepts at different granularity (e.g., person vs. politician vs. president), and (ii) contextual subtlety on fine-grained types (e.g., Boston vs. Detroit could be two sports teams, instead of two cities). 
Most existing methods utilize weak or distant supervision to automatically generate training data for the FET tasks. There are three major approaches to obtaining weakly-labeled training data to tackle these challenges. 
The first is to automatically match the mentions in text with the concepts in some existing \emph{knowledge bases} (e.g., Wikipedia) \cite{Weld_Ling2012}. The typical workflow is to first detect entity mentions from a corpus, map these mentions to knowledge base (KB) entities of target types, and then leverage the confidently mapped types as pseudo-labeled data to infer the final type.
The second is to directly obtain the \emph{head words} of nominal mentions as its fine-grained type \cite{choi-etal-2018-ultra}. This approach leverages the head words of the entity mention to consolidate context-aware types derived from the KB matching. However, both approaches suffer from label sparsity and context-agnostic problems, resulting in the inability to generate high-quality training data for FET. 

The third approach is to \emph{probe the pre-trained language models} through the use of masked patterns and entailment templates. This method enables the use of PLM as a knowledge base for weak supervision. Leveraging masked language model (MLM) prompting to generate rich and context-aware weak supervisions for fine-grained entity typing is a recent trend, aiming to leverage a PLM's contextual and  ``common-sense'' knowledge learned at training to reduce expensive human labor \cite{dai-etal-2021-ultra, li-etal-2022-lite}.

Given a sentence that contains a mention, one can use a ``[MASK]'' token to replace the entity mention span to generate candidate entity types. 
However, such a method is unguided which may result in the generation of tokens that are inadequate for entity typing (e.g., \{This, That, Him, Me, It\} for “Wrigley Field'' in $S_1$).

Alternatively, we can seek to extract hypernyms of the entity mention span of interest by inserting a short phrase that contains a ``[MASK]'' token (e.g., exploring the idea of Hearst Patterns \cite{hearst-1992-automatic}). This method conducts labeling in a more context-aware manner and greatly enriches the fine-grained types labeled for each mention. 

This process, however, still has the potential to generate tokens unsuitable for typing (e.g., \{Team, Thing\} for “Wrigley Field'' in $S_1$) or a mixture of rough and fine-grained types (e.g., \{Location, Building, Stadium\}). 
The difficulty cannot be resolved automatically due to the lack of hierarchical knowledge of the generated tokens/types.

With the emergence of large language model (LLM) (e.g., ChatGPT \cite{openai2023gpt4}), it is appealing to directly apply it to tackle these challenges. 
However, due to the lack of knowledge structures, an LLM may generate entity types at a too coarse or too fine level, or fail to commit to the right one from numerous fine-grained candidates.

In this study, we envision that an ontology structure, which provides a semantics-rich, hierarchical structure, may help select the best results generated by multiple PLM models.
We propose a \underline{annotation-free}, \underline{ontology-guided}, \underline{fine-grained} entity typing (FET) method, \our, that leverages an input ontology structure and the power of MLM prompting and Natural Language Inference (NLI).
We first ensemble multiple Hearst patterns to perform MLM prompting, reducing the noise in the initial candidate type generation. 
Since the generated candidate labels for a given mention are likely a mixture of fine and coarse-grained labels, or tokens unsuitable for typing, we propose to automatically match the generated candidate labels to a coarse-grained type in our type ontology and then rank and select a coarse-grained type with a pre-trained entailment model under the local context. 
Based on the same principle of entailment score-based type selection,
this type resolution process progresses deeper to finer levels,  following the type ontology, until the finest possible label can be consolidated.

\vspace{0.2ex}
\noindent
\textbf{Example 2.} 
For sentence $S_1$ in Ex.\ 1, candidate type generation (Step 1) ensembles prompting results of multiple Hearst patterns and generates a set of candidate labels: \{Stadium, Venues, Location, Game\} for ``Wrigley Field'' (Fig.\ \ref{fig:candTypeGen}).
Using a given ontology structure (Fig.\ \ref{fig:typeOntology}), this set of types is first resolved to the course-grained type ``Location'' via the assistance of a pre-trained entailment model and the local contextual information (Fig.\ \ref{fig:typeAlign}).
Note without the type structure, ``Stadium'' and ``Venue'' are rivals to ``Location''; but with the structure,  they become its supporters since both are fine-grained ``Location''.
With the same principle, the type resolution proceeds deeper to finer-grained levels, along the type ontology, from ``Location'' to  ``Building'' and further down to ``Stadium'' for ``Wrigley Field'', leading to the most accurate fine-grained type (Fig.\ \ref{fig:typeRes}).

\vspace{0.3ex}
\noindent
Our contributions are summarized as follows.
\vspace{-0.2ex}
\leftmargini=12pt
\begin{enumerate}
\parskip 0.2ex
\item A fully annotation-free, ontology-guided, fine-grained typing method, \our, is proposed
\item \our 
improves fine-grained entity typing (FET) by leveraging candidate labels generated and refined with three information sources: (i) pre-trained language models, (ii) a fine-grained type ontology, and (iii) head words
\item Experiments on the Ontonotes, FIGER, and NYT datasets \cite{Gillick2014} using their associated ontological structures show that \our clearly outperforms existing zero-shot named entity typing methods as well as ChatGPT, and even rivals supervised methods.
Our error analysis shows that refinement of ontology structures will further improve fine-grained entity typing.
\end{enumerate}
\vspace{-0.25mm} 

\section{Related Work}
Fine-grained entity typing benefits various downstream tasks and has received extensive attention in natural language research. Recent studies focus on different contexts from the phrase level \cite{yao2013universal} to considering specific entity mentions in the sentence or document level \cite{Weld_Ling2012, gupta-etal-2017-entity, choi-etal-2018-ultra}. Entity typing has been generally studied under supervised learning settings with significant human annotated data \cite{Weld_Ling2012, yosef-etal-2012-hyena, choi-etal-2018-ultra, dai-etal-2021-ultra}.

\smallskip
\noindent
\textbf{Minimally Annotated Fine-Grained Entity Typing Methods.} 
Some recent studies (e.g., \cite{zhang-etal-2020-empower, dai-etal-2021-ultra, liu2021survey, li-etal-2022-lite, Li2023UltraFineET, jiang-etal-2023-recall, Xu2023SemisupervisedLF, zhang-etal-2023-denoising}) leverage cross-encoders, pre-trained language models and prompting templates to obtain knowledge for entity mentions in given sentences.
\cite{dai-etal-2021-ultra} improves ultra-fine entity typing with a BERT Masked Language Model (MLM). 
\cite{li-etal-2022-lite} instead improves ultra-fine entity typing by treating the task of predicting an entity type as an NLI task.
Similar to \our, ChemNER \cite{wang2021chemner} leverages a type ontology structure to guide fine-grained Chemistry entity recognition. 
NFETC-SSL \cite{Xu2023SemisupervisedLF} proposes a semi-supervised learning method with mixed label smoothing and pseudo labeling. MCCE \cite{jiang-etal-2023-recall} proposes to first prune the large type set and generate K candidates. Then, MCCE's novel model concurrently encodes and scores these K candidates as final type labels. DenoiseFET \cite{Li2023UltraFineET} utilizes pre-trained label embeddings to group the given set of type labels into semantic domains. A fine-tuned UFET model is utilized to predict initial labels. Finally, the semantic domains as guidance to infer missing labels and remove conflicting labels.
These recent methods leverage human annotations and seek to augment SOTA models with prior knowledge acquired through head words, ontology structures and pre-trained embeddings. \our seeks to perform FET without the use of any human annotations by amalgamating multiple MLM prompts and NLI results to reduce noises in candidate type generation. Finally, we also utilize the fine-grained type ontology structure as guidance to progressively resolve candidate labels from coarse to fine under the local context.

\smallskip
\noindent
\textbf{Distant Supervision via Knowledge Bases.} 
To handle difficulties to acquire human annotation, the zero-shot learning setting has been introduced for named entity typing \cite{xia-etal-2018-zero}. 
Several studies \cite{yuan2018otyper,zhou-etal-2018-zero} address the problems by grounding the mentions with Knowledge Base (i.e. Wikipedia) entries from the assembled related pages. These methods achieve good performance but still require significant human resources to construct effective knowledge bases which can be difficult to obtain for emerging domains. 

\smallskip
\noindent
\textbf{Transfer Learning Methods.} 
Other studies explore learning pre-trained semantic word-level embeddings from Knowledge Bases and seen types \cite{ren-etal-2016-afet, ma-etal-2016-label} or extracting raw embeddings without auxiliary information and utilize end-to-end neural networks \cite{zhang-etal-2020-mzet}. However, these methods still suffer from low accuracy and inefficiency in zero-shot settings. As a result, \our turns to the weak supervision by exploiting pre-trained language models (i.e. BERT \cite{devlin-etal-2019-bert}) as a knowledge base due to their substantial knowledge in language understanding learned by training on massive text collections.

\section{Methodology}

We propose \our, an annotation-free, ontology-guided, fine-grained entity typing method using pre-trained language models and a fine-grained ontology structure. Given an input sentence and a set of pre-identified mentions in the sentence,
\our consists of the following steps: (1) generating a set of candidate labels for each input mention with both head word parsing and an ensemble of MLM prompting (Fig.\ \ref{fig:candTypeGen});
(2) resolving the coarse-grained types by matching and ranking the generated labels to the type ontology using an entailment model (Fig.\ \ref{fig:typeAlign}); and (3) progressively refining the fine-grained types along the type ontology following the principle of entailment score-based type selection (Fig.\ \ref{fig:typeRes}).
We utilize the inherent structure of the fine-grained type ontology and a pre-trained entailment model (RoBERTa model pre-trained on the MNLI dataset \cite{roberta}) to guide our fine-grained entity typing.

\subsection{Problem Definition}

The input to our proposed \our framework is a text corpus $D$ and a fine-grained type ontology $O$. In this study, we assume our input text corpus $D$ includes a set of pre-identified entity mentions. An entity mention, $e$, is a token span in the text document that refers to a real-world entity.  Given a sentence $S$ and a parsed entity mention $e \in S$, the \textbf{fine-grained entity typing} (FET) task is to identify one or more types $t$ from the label space $T$ (provided in a structured ontology $G$) for the entity mention $e$. 

As an example of our FET task, in the sentence $S_1$ of Ex.\ 1, the entity mention to be typed is $e_1$: ``\textit{Wrigley Field}''. 
It should be labeled progressively deeper as ``Location$\rightarrow$Building$\rightarrow$Stadium'' as opposed to other labels like ``Organization'', ``Person'', or ``School''. 
\subsubsection{Fine-Grained Type Ontology Structure.}
The structure of the type ontology is fundamental to the \our algorithm. In this study, we utilize zero human effort to construct our type ontologies. We leverage the fine-grained type ontologies provided in the OntoNotes and FIGER datasets. Extending these ontologies to new text corpora can be achieved with minimal human efforts through taxonomy completion and refinement methods\cite{Shen2020TaxoExpanST, Jiang2022TaxoEnrichST, Jiang2023ASV, Arous2023TaxoCompleteST}. Furthermore, existing automatic ontology constructions methods \cite{velardi2013ontolearn, wang2017short, wu2012probase} build ontologies also based on “is-A” relations (e.g., a “Canada” is a “Country”) which minimize the need for human-annotated ontologies even in emerging domains.

\begin{figure}[ht]
    \centering
    \includegraphics[width=0.8\columnwidth]{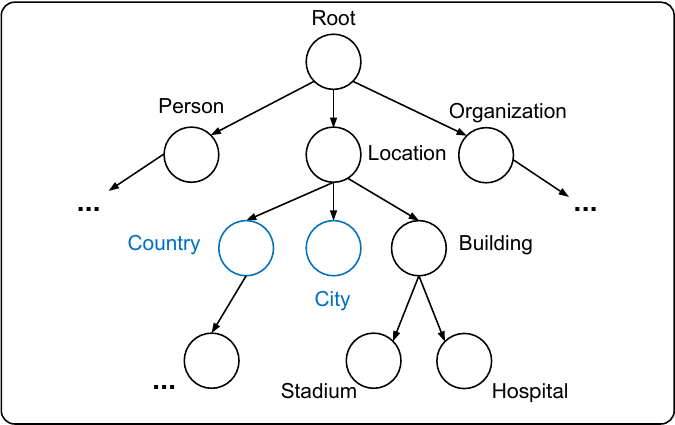}
    \caption{OntoNotes Type Ontology}
    \label{fig:typeOntology}
    \vspace{-0.75cm}
\end{figure}

\begin{definition}[Fine-Grained Type Ontology]
\our's Fine-Grained Type Ontologies are structured as a strict tree imposing an ``is-a'' type hierarchy stemming from a ``root'' concept. The ``root'' concept's children consist of a handful of coarse-grained types including but not limited to: ``Organization'', ``Person'', or ``Location''. In addition, the ontology follows the following structural constraints: (1) Each type has a singular parent type; (2) each type (except for the leaf node) can have a number of children; and (3) each type is connected by a directional edge indicating a hypernym-hyponym type relationship between the parent and child nodes.
\end{definition}
It is important to recognize that a hypernym-hyponym or ``is-a'' relationship is critical for guiding \our's final fine-grained type selection. Note that in the given OntoNotes type ontology (Fig.\ \ref{fig:typeOntology}), City and Country are sibling types since they share the same parent type Location. Although ``City'' can be connected to ``Country'' by a ``is-in'' type relationship, \our's input ontologies organize these entity types as siblings due to their hypernym-hyponym relationship to ``Location''. By consolidating hypernyms of the entity mention, we can identify the high-level type in the ontology that accurately provides a partial label for the entity mention of interest. Thus, we approach our entity typing process in a hierarchical manner and refine from a coarse-grained level down to the most accurate fine-grained level for the mention of interest. 

Please note that an ontology structure can be extended to include
``is-in'', ``is-part-of'' and ``is-property-of'' relationships. 
Fine-grained entity typing using such an extended structure is left to future work.

\subsection{Candidate Type Generation}
To generate a set of candidate entity types for each input mention, we leverage two techniques: (1) head word parsing, and (2) ensemble of hypernyms derived from MLM prompting. 

Head words and hypernyms can serve as powerful context-aware type indicators that can be leveraged as weak supervision sources for entity typing \cite{choi-etal-2018-ultra, dai-etal-2021-ultra}. We use them to select a set of candidate types to guide our fine-grained type refinement. 
We first generate and parse the head words for a given entity mention, and then generate candidate types with the use of ensembled MLM prompting\footnote{If there is no head word, our algorithm generates candidate labels using only MLM prompting.}. The generated candidates are used as input to the following steps of ontology-guided type resolution.

\subsubsection{Head Word Parsing.}
As discussed in \cite{choi-etal-2018-ultra}, the input text often contains cues that explicitly match a mention to its type. These cues are often in the form of the mention’s head word. Thus, given the pre-identified entity mentions in the input sentence, we first identify the head word of the input mention. We utilize the Stanford Dependency Parser \cite{chen-manning-2014-fast} to extract the head word of the entity that we are interested in typing. Considering the sentence ``\emph{Governor Arnold Schwarzenegger gives a speech at Mission Serve's service project on Veterans Day 2010}'', 
the entity mention ``\textit{Governor Arnold Schwarzenegger}'' can be associated with a few types
(e.g., actor, father, and governor). However, given the head word, one can easily consolidate it to the right one: ``governor''.  Thus, \our leverages the head word of the input entity, if any, to select an initial context-sensitive type, which may guide the selection of the subsequent fine-grained type.

\subsubsection{Ensembled MLM Prompting.}
\begin{figure}[ht]
    \vspace{-0.3cm}
    \centering
    \includegraphics[width=0.8\columnwidth]{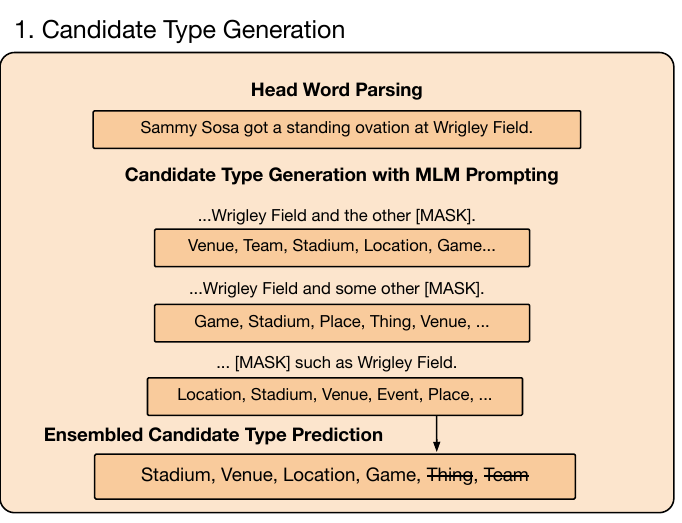}
    \caption{Candidate Type Generation}
    \label{fig:candTypeGen}
    \vspace{-0.3cm}
\end{figure}

While head words can provide strong type indicators, they do not always provide sufficient information to consolidate a high-level type. In some cases, head words (e.g., ``Red Sox star'') cannot directly provide accurate type information as they are not directly present in the input type ontology. 
Furthermore, head words are not always available in the input sentences. Thus, with the parsed entity mention span as input, we propose to leverage context-aware hypernyms as initial type candidates for the target mention. 
With an ensembled cloze prompting method, \our generates candidate types of the mention with masked language models and performs an initial high-level typing on the input mention. Specifically, we leverage the BERT model \cite{devlin-etal-2019-bert} and artificial Hearst patterns \cite{hearst-1992-automatic} to generate context-aware hypernyms that serve as candidate types for the mentions. 
We first modify the input sentence by inserting a Hearst pattern and [MASK] token into the sentence. Then we use the BERT to generate candidate types for the target mention under the local context. For example, in Fig.\ \ref{fig:candTypeGen}, we first insert Hearst patterns such as ``and the other [MASK]'' in the input sentence and then use the BERT MLM model to generate candidate types such as ``Venue'', ``Team'', and ``Stadium''.

We evaluated the quality of hypernyms generated with direct masked prompting and the 44 patterns proposed in \cite{seitner-etal-2016-large} on the Ontonotes Development Dataset. When generating hypernyms, we expect high-quality candidates to be semantically equivalent to concepts contained in our input type ontology. 

Unfortunately, we found that direct masked prompting resulted in tokens that were indefinite and unsuitable to serve as candidate types. 
Based on our observation, the four Hearst patterns in Table \ref{tab:Hearst} provide the highest quality hypernyms under a simple direct matching to types in the OntoNotes ontology. 

Nevertheless, based on the syntax and grammar of the sentence, hearst patterns can generate tokens that are unsuitable to serve as fine-grained entity types. For example, with a single prompt, the MLM can generate ``Famous'', ``Actor'', ``Celebrity'' and ``Person'' as the most probable words, where ``Famous'' is unsuitable to serve as a fine-grained entity type. 
To reduce the noises caused by the use of a single Hearst pattern, we ensemble $n$ Hearst Patterns to consolidate the most commonly generated candidate types. For each pattern in the pattern list $P$, we collect the top $k$ most probable tokens from the probability distribution predicted by the BERT MLM. Then, we aggregate the tokens and identify the set of candidates that have the largest overlap. 
We perform the voting ensemble as follows:
\begin{equation}
y = count_{\left( m \right) }\{H_1(x), H_2(x), ..., H_n(x)\}
\end{equation}
where $H_n(x)$ is the candidate type generated by the $n^{th}$ hearst pattern and $count_{\left( m \right)}$ is the function that selects all candidates generated at least $m$ times.   
In our experiments, we take $m= \lfloor \frac{n}{2} \rfloor + 1 $.
We observe that the number of Hearst patterns employed is not sensitive as ensembling ensures the most confident candidates retain.

\noindent
\textbf{Example 3}: As shown in Fig.\ \ref{fig:candTypeGen}, by ensembling the results from prompting with several Hearst patterns, the quality types for $e_1$ ``Stadium, Venue, Location, Game'' retain but the noisy types ``Thing'' and ``Team'' are removed.

\begin{table}[]
\centering
\small
\begin{tabular}{llll}
\hline
    \textbf{Masked Prompt Pattern} & \textbf{Prec} & \textbf{Rec} & \textbf{F1}  \\ \hline
    [MASK] & 11.8 & 5.2 & 7.2  \\\relax
    [MASK] such as & 53.3 & 72.4 & 61.4  \\ 
    such [MASK] as & 47.9 & 68.7 & 56.5  \\ 
    and some other [MASK] & 48.8 & 66.6 & 56.4  \\ 
    and the other [MASK] & 47.6 & 68.3 & 56.1  \\ \hline
\end{tabular}
\caption{Performance of direct masked token prompting \& Hearst patterns with simple type mapping on Ontonotes Dataset.}
\label{tab:Hearst}
\vspace{-0.75cm}
\end{table}

\subsection{High-Level Type Resolution}
Given the generated candidate types and head words for each mention in the sentence, we seek to resolve the concrete type for this candidate type set at the high levels of the type ontology. Specifically, we first align the generated candidates to several high-level types in the type ontology, and then select the most accurate high-level types with a pre-trained entailment language model.

\subsubsection{Candidate Type Alignment.}
Following the previous step of candidate type generation, we combine the generated candidates from both the parsed head words and the ensembled cloze prompting to form a candidate type set. These candidates are generally noisy and may not exist directly in our type ontology. However, we observe that the generated candidates will usually cluster closely around a high-level concept in the ontology. Thus, we perform our high-level type alignment with a cosine-similarity-based matching.

We use Word2Vec\footnote{https://code.google.com/archive/p/word2vec/} \cite{NIPS2013_9aa42b31} embeddings to construct our type embeddings for the cosine-similarity-based type alignment. 
We construct a verbalizer by selecting at least $l$ semantically related words for each coarse type. 
For a high-level type of ``Organization'', we might include seed types such as ``Corporation'', ``University'', ``Firm'', ``Business'', and ``Government'' in its verbalizer. This verbalizer is then utilized to systematically map the MLM hypernym prediction to the most relevant type. In our experiments, we provide at least five seed types $S$ for each type node $c$ in the first level of the type ontology. 
Increasing the number of seed types increases the coverage and confidence of the verbalizer. 
Since the most commonly generated hypernyms for each concept are in the input ontologies, \our is not sensitive to the number of seed types collected.
With the seed types, we construct a node embedding $N$ by taking the average of word embeddings from both the first-level type and its corresponding seed types,
\begin{equation}
N = \frac{\sum_{s_i \in S}emb(s_i) + emb(c)}{|S|+1}.
\end{equation}
where $emb(\cdot)$ indicates the Word2Vec embeddings. 

Then we rank each generated candidate type to a first-level type on our type ontology by calculating the cosine similarity between the embeddings of the generated candidate labels $b$ and that of the first-level types $T$,
\begin{equation}
score(b,T) \myeq{rank} cosine(emb(b), emb(T)).
\end{equation}
Finally, the first-level type that has the highest similarity is selected as the aligned high-level type. 

\begin{figure}[ht]
    \vspace{-0.3cm}
    \centering
    \includegraphics[width=0.8\columnwidth]{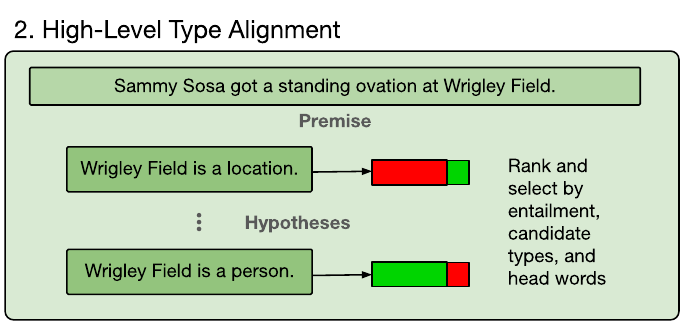}
    \caption{Candidate Type Generation}
    \label{fig:typeAlign}
    \vspace{-0.3cm}
\end{figure}
\noindent
\textbf{Example 4}: In Figures \ref{fig:candTypeGen} and \ref{fig:typeAlign}, the consolidated candidate labels ``Location'', ``Stadium'', ``Venue'', and ``Game'' are closely related to the high-level type ``Location'' in our ontology. By performing the cosine-similarity-based matching, ``Location'' is identified as $e_1$'s high-level type over ``Organization'' or ``Person''. Thus, we select the most similar high-level type given by the head word, generated candidate labels, and the entailment model.

\subsubsection{High-Level Type Selection}
After the previous step of candidate type alignment, we obtain several high-level types for each entity mention in the sentence. Given these types, we seek to select the most accurate high-level type for each entity mention under the local context.
We observe that the task of selecting the most appropriate entity type can be viewed as a Natural Language Inference (NLI) task. Thus, we treat the input sentence as the premise in NLI and generate the hypothesis using a pre-defined declarative template. 
To resolve the type of the input mention, we use the template: ``\texttt{In this sentence, [MENTION] is a [TYPE].}''
We then rank each type in the first level of the ontology with the entailment score from a RoBERTa NLI model \cite{roberta}.

\smallskip
\noindent
\textbf{Example 5}: In Fig.\ \ref{fig:typeAlign}, we align a majority of the generated candidate type set to the Location seed types. The NLI model further ranks Location over Organization or Person. By utilizing the information in conjunction, $e_1$ is solidly aligned to the high-level type ``Location''. 

\subsection{Fine-Grained Type Resolution}

Given the high-level types of the entity mentions, \our further leverages the ontology structure to progressively resolve the fine-grained label. Following the same principle of entailment-based type selection for the high-level types, we utilize the entailment model \cite{roberta} to compute the entailment score, $\sigma_{entail}$. Then, \our can automatically select the most accurate fine-grained entity types along our type ontology. Specifically, we first examine the child types of the previously determined higher-level types and then select the child type with the highest ranked score as the fine-grained type. 

In addition to the entailment model, we also utilize the candidate type set to refine our fine-grained type selection. If the parsed head word is present in our type ontology, the entailment scores of that parent and its child types are weighted higher through $\sigma_{cand}$. Similarly, if the generated candidate types are in our type ontology, the entailment scores of their parents and children are also weighted higher through $\sigma_{cand}$. Finally, we select the fine-grained type for each mention with the highest-ranked score.

We calculate the ranking score for the entities at the current level of the ontology as follows:
\begin{equation}
rank(type) = \sigma_{entail} + \sigma_{cand}
\end{equation}
We first leverage our NLI pre-trained model \cite{roberta} to find the entailment score $\sigma_{entail}$ for each entity type. Then if a type in the candidate type set is a descendent to the entity type, we add a weight $\sigma_{cand}$.
We repeat this entailment-based selection process recursively along the type ontology to select the best fine-grained type. 

\smallskip
\noindent
\textbf{Definition} (Entity Type Granularity Parameter $\theta$): We assume there is a scalar of $\theta$ indicating how granular or specific the final selected entity type should be. The smaller $\theta$ is, the more granular entity types are consolidated as the final one.

Thus, if the child types do not have a sufficient gain of at least $\theta$ in ranking score over the parent type at a certain level, we stop the recursion and select the parent type as the final fine-grained type.
We conduct a parameter study to explore the sensitivity of \our to the parameter $\theta$ in Section \ref{sec:paramSens}.

\begin{figure}[ht]
    \vspace{-0.3cm}
    \centering
    \includegraphics[width=0.8\columnwidth]{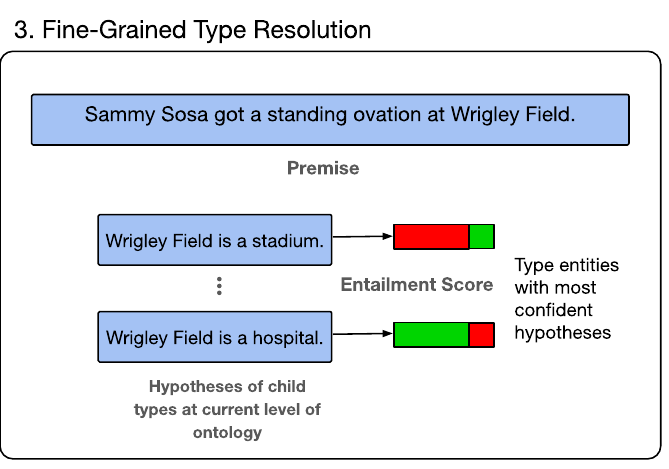}
    \caption{Fine-Grained Type Refinement}
    \label{fig:typeRes}
    \vspace{-0.3cm}
\end{figure}

\noindent
\textbf{Example 6}: 
In Fig.\ \ref{fig:typeRes}, we consider all descendent types of ``Location'' as potential fine-grained entity types for $e_1$. To begin, \our generates the hypotheses and ranks all child types of ``Location''. With the resultant entailment score rankings of these sibling types, we consolidate and select ``Building'' as the highest-ranked type. 
Then, a similar process is done at a deeper level of the ontology to select the final type ``Stadium''.

\section{Experiments}
\begin{table}[ht]
\centering
\small
\begin{tabular}{llll}
\hline
    \textbf{Datasets} & \textbf{Ontonotes} & \textbf{FIGER} & \textbf{NYT} \\ \hline
    \# of Types & 89 & 113 & 125\\ 
    \# of Documents & 300k & 3.1M & 295k\\ 
    \# of Entity Mentions & 242K & 2.7M & 1.18M\\
    \# of Train Mentions & 223K & 2.69M & 701K\\
    \# of Test Mentions & 8,963 & 563 & 1,010\\ \hline
\end{tabular}
\caption{Dataset Statistics}
\label{tab:Datasets}
\vspace{-1cm}
\end{table}

\begin{table*}[]
\footnotesize
\begin{tabular}{ll|lll|lll|lll}
\hline
    \multirow{2}{*}{\textbf{Settings}}& \multirow{2}{*}{\textbf{Model}} & \multicolumn{3}{c|}{\textbf{NYT}} & \multicolumn{3}{c|}{\textbf{FIGER}} & \multicolumn{3}{c}{\textbf{Ontonotes}} \\
     &  & \textbf{Acc} & \textbf{Mi-F1} & \textbf{Ma-F1} & \textbf{Acc} & \textbf{Mi-F1} & \textbf{Ma-F1} & \textbf{Acc} & \textbf{Mi-F1} & \textbf{Ma-F1} \\ \hline
    \multirow{2}{*}{\textbf{Weak Supervision}} & UFET \cite{choi-etal-2018-ultra} & - & - & - & - & - & -  & 59.5 & 71.8 & 76.8  \\ 
    \multirow{2}{*}{\textbf{with Human Annotations}} & BERT-MLMET \cite{dai-etal-2021-ultra} & - & - & - & - & - & - & 67.44 & 80.35 & 85.44  \\ 
    & LITE \cite{li-etal-2022-lite} & - & - & - & 66.2 & 74.7 & 80.1 & 68.2 & 81.4 & 86.6  \\ 
    & NFETC-SSL \cite{NFETCSSL2023} & - & - & - & 71.2 & 80.2 & 81.9 & 64.4 & 74.3 & 79.7  \\ \hline
    \multirow{2}{*}{\textbf{Distant Supervision via KBs}} & DZET \cite{obeidat-etal-2019-description} & 27.3 & 53.1 & 51.6 & 28.5 & 56.0 & 55.1  & 23.1 & 28.1 & 27.6  \\ 
    & ZOE \cite{zhou-etal-2018-zero} & 62.1 & 73.7 & 76.9 & \textbf{58.8} & \textbf{71.3} & 74.8  & 50.7 & 60.8 & 66.9  \\ \hline
    \multirow{2}{*}{\textbf{Transfer Learning}} & OTyper \cite{yuan2018otyper} & 46.4 & 65.7 & 67.3 & 47.2 & 67.2 & 69.1  & 31.8 & 36.0 & 39.1  \\ 
    & MZET \cite{zhang-etal-2020-mzet} & 30.7 & 58.2 & 56.7 & 31.9 & 57.9 & 55.5  & 33.7 & 43.7 & 42.3  \\ \hline
    & ChatGPT Prompt 1\cite{openai2023gpt4} & 47.3 & 59.1 & 54.3 & 51.7 & 65.3 & 58.3 & 27.7 & 37.5 & 32.6   \\ 
    & ChatGPT Prompt 2 \cite{openai2023gpt4} & 45.1 & 64.0 & 61.9 & 52.3 & 67.8 & 61.4 & 31.3 & 41.3 & 35.9 \\ 
    & ChatGPT Prompt 3 \cite{openai2023gpt4} & 44.8 & 56.9 & 52.1 & 49.9 & 61.1 & 55.8 & 24.7 & 33.8 & 29.4 \\ 
    & Gemma \cite{gemmateam2024gemma} & 44.8 & 56.9 & 52.1 & 49.9 & 61.1 & 55.8 & 24.7 & 33.8 & 29.4   \\ 
    & LLaMA 2 \cite{touvron2023llama} & 43.2 & 55.6 & 51.7 & 48.5 & 59.7 & 54.4 & 24.1 & 33.2 & 28.8 \\ 
    \textbf{Annotation-Free} & \textbf{\our} + Original Ontology & \textbf{69.6} & \textbf{78.4} & \textbf{82.8} & 49.1 & 67.4 & \textbf{75.1}  & \textbf{65.7} & \textbf{73.4} & \textbf{81.5}  \\ 
    & \textbf{\our} + Modified Ontology & - & - & - & 51.1 & 68.9 & \textbf{77.2} & - & - & - \\ \hline
\end{tabular}
 \caption{Results (\%) on Three Test Sets (Some slots in the benchmarked methods marked "-" due to no fully annotated training data).}
 \vspace{-0.5cm}
\label{tab:results}

\end{table*}

\subsection{Datasets}

We compare the performance of \our and several baseline methods on three benchmark FET datasets: NYT, Ontonotes \cite{Gillick2014} and FIGER \cite{Weld_Ling2012}. The basic statistics of the datasets are shown in Table \ref{tab:Datasets}. For the OntoNotes and FIGER datasets, we leverage the included type ontologies while the NYT dataset leverages the input FIGER ontology. All NER test sets are annotated using the ontologies as a set of type labels. Thus, each entity mention is labeled with a fine-grained label represented as a path within the ontology. The ontologies have a maximum depth of three and contain four to six high-level types (e.g., Location, Person, and Organization).

\subsection{Baselines}
\our is a FET method that does not require human annotation as supervision. We compare \our with four weakly supervised and nine zero-shot FET methods utilizing the evaluation metrics detailed in Appendix \ref{sec:evalMetrics}. 

For each baseline method, we utilize the default parameters as detailed in their studies. Furthermore, we conduct all experiments on a single NVIDIA RTX A6000 GPU. Finally, we evaluate generative LLMs on FET using Gemma 2b model\cite{gemmateam2024gemma}, Llama 2 7b model\cite{touvron2023llama} and OpenAI “gpt-3.5-turbo”\cite{openai2023gpt4}. \our leverages a pre-trained BERT \cite{devlin-etal-2019-bert} (BERT-base, uncased) and pre-trained RoBERTa fine-tuned on the MNLI dataset \cite{roberta} available in the HuggingFace Library. In addition, \our utilizes Word2Vec\footnote{https://code.google.com/archive/p/word2vec/} \cite{NIPS2013_9aa42b31} embeddings to construct our type embeddings. Finally, we conduct parameter and ablation studies listed in Sections \ref{sec:paramSens} and \ref{sec:abla} respectively.

\vspace{0.3ex}
\noindent
\textbf{Weak Supervision with Human Annotations.}

\noindent 
{\sf UFET} \cite{choi-etal-2018-ultra}: This weakly supervised baseline is a model that predicts open types and is trained using a multitask objective that pools head-word supervision with supervision from entity linking.

\noindent 
{\sf BERT MLMET} \cite{dai-etal-2021-ultra}: This weakly supervised baseline fine-tunes a BERT-based model using human annotations, supervision from head words and mention hypernyms generated from Hearst patterns.

\noindent 
{\sf LITE} \cite{li-etal-2022-lite}: This weakly supervised baseline is a fine-tuned MNLI model leveraged to rank ultra-fine entity types.

\noindent
{\sf NFETC-SSL} \cite{NFETCSSL2023}: This weakly supervised baseline proposes a semi-supervised learning method with mixed label smoothing and pseudo labeling for fine-grained entity typing.

\vspace{0.3ex}
\noindent
\textbf{Distant Supervision from Knowledge Bases.}

\noindent 
{\sf ZOE} \cite{zhou-etal-2018-zero}: This zero-shot baseline leverages a type taxonomy defined as Boolean functions of Freebase types and grounds a given entity mention to the type-compatible Wikipedia entries to infer the target mention’s types.

\noindent 
{\sf DZET} \cite{obeidat-etal-2019-description}: This zero-shot baseline utilizes the type description available from Wikipedia to build a distributed semantic representation of the types and aligns the target entity mention type representations onto the known types.

\vspace{0.3ex}
\noindent
\textbf{Transfer Learning Methods.}

\noindent 
{\sf OTyper} \cite{yuan2018otyper}: This zero-shot baseline is a neural network trained on a limited training dataset. The model is evaluated on the Open Named Entity Typing task, which is FET where the set of target types is not unknown.

\noindent 
{\sf MZET} \cite{zhang-etal-2020-mzet}: This zero-shot baseline leverages the semantic meaning and the hierarchical structure into the type representation. The method leverages the knowledge from seen types to label the zero-shot types through the use of a memory component.

\vspace{0.3ex}
\noindent 
{\sf ChatGPT} \cite{openai2023gpt4}: This zero-shot, annotation-free baseline leverages a generative large language model as a knowledge source for entity typing. We evaluate ChatGPT on three different prompts to mitigate possible prompt construction-based errors. 

Prompt1: \emph{“Return the fine-grained entity types of the given entity mentioned in the sentences below. Be concise and ONLY utilize the types from this list of possible entity types.
[entity types] \{entity types\} [sentence] \{sentence\} [entity mention] \{entity mention\}”}.

Prompt2: \emph{“Select a single label from the following list that best serves as a hyponym for the entity mention. [labels] \{entity types\} [sentence] \{sentence\} [entity mention] \{entity mention\}"}

Prompt3: \emph{"Select the single most fine-grained entity type from the following list for the given entity mention. [entity types] \{entity types\} [sentence] \{sentence\} [entity mention] \{entity mention\}"}

We discuss the limitations of leveraging ChatGPT as a fine-grained entity typing method in Section \ref{sec:LLMAnalysis}.

\vspace{0.3ex}
\noindent 
{\sf Gemma} \cite{gemmateam2024gemma}: This zero-shot, annotation-free baseline leverages a generative large language model as a knowledge source for entity typing. We utilize Prompt1 for evaluation on the FET task.

\vspace{0.3ex}
\noindent 
{\sf LLaMA 2} \cite{touvron2023llama}: This zero-shot, annotation-free baseline leverages a generative large language model as a knowledge source for entity typing. We utilize Prompt1 for evaluation on the FET task.

\subsection{Main Results}

Table \ref{tab:results} shows our results on the test set of NYT, FIGER and Ontonotes, especially comparing with the notable exising method ZOE.
On the NYT dataset, \our achieves the best zero-shot performance on this dataset. 
It achieves 5.9 absolute Ma-F1 improvement over the state-of-the-art zero-shot fine-grained entity typing method ZOE. 
While on the Ontonotes dataset, \our achieves the best zero-shot performance on this dataset while trailing the best supervised method by 3.94 Ma-F1 points. \our achieves 14.6 absolute Ma-F1 improvement over the state-of-the-art zero-shot fine-grained entity typing method ZOE. 

A performance comparison between \our and ZOE demonstrates the benefit of leveraging the knowledge embedded in pre-trained language models as a form of minimal supervision to identify entity mention types. ZOE leverages a type ontology and maps a given mention to type-compatible Wikipedia Entries. As a result, ZOE relies on surface-level information from the mention string. OntoType ensembles contextual information from various PLMs to consolidate the final entity type. Given a sentence: “\emph{The biggest cause for concern for McGuff is the bruised hamstring Regina Rogers suffered against (Utah) last Saturday}”, ZOE incorrectly utilizes the surface string to label “Utah” as a location. With PLMs, OntoType recognizes ``Team'' or ``Opponent'' as candidates and finally consolidates to  ``Sports Team''. 
\begin{table}[ht]
\footnotesize
\begin{tabular}{p{1.5cm}|p{5.6cm}}
\hline
    \textbf{MZET} & \RaggedRight \textbf{US President Joe Biden} \textbf{\textcolor{teal}{\textbackslash Person\textbackslash Politician}} was one of many foreign leaders to speak with President Zelensky, and he "pledged to continue providing \textbf{Ukraine} \textbf{\textcolor{teal}{\textbackslash Location}} with the support needed to defend itself, including advanced air defence systems", \textbf{the White House} \textbf{\textcolor{teal}{\textbackslash Location\textbackslash Building}} said. \\ \hline
    \textbf{ZOE} & \RaggedRight \textbf{US President Joe Biden} \textbf{\textcolor{violet}{\textbackslash Person\textbackslash Politician}} was one of many foreign leaders to speak with President Zelensky, and he "pledged to continue providing \textbf{Ukraine} \textbf{\textcolor{violet}{\textbackslash Location\textbackslash Country}} with the support needed to defend itself, including advanced air defence systems", the \textbf{White House} \textbf{\textcolor{violet}{\textbackslash Location\textbackslash Building}} said. \\ \hline 
    \textbf{\our} & \RaggedRight \textbf{US President Joe Biden} \textbf{\textcolor{purple}{\textbackslash Person\textbackslash Politician\textbackslash President}} was one of many foreign leaders to speak with President Zelensky, and he "pledged to continue providing \textbf{Ukraine} \textbf{\textcolor{purple}{\textbackslash Location\textbackslash Country}} with the support needed to defend itself, including advanced air defence systems", \textbf{the White House} \textbf{\textcolor{purple}{\textbackslash Organization\textbackslash Government}} said. \\ \hline  
\end{tabular}
\caption{Type predictions (in color) on three entity mentions (in bold): 
ONTOTYPE vs.\ two other Zero-Shot FET methods.}
\label{tab:Case1}
\vspace{-0.75cm}
\end{table}

On the FIGER dataset, \our achieves 0.3 absolute Macro-F1 improvement over state-of-the-art zero-shot fine-grained entity typing method ZOE \cite{zhou-etal-2018-zero}.
However, our method trails ZOE in strict accuracy and Micro-F1. In the FIGER dataset, predictions are made based on both the surface-level information and the contextual information in the sentence. The major advantage of \our is to accurately capture the contextual information to provide a fine-grained entity type. 
However, \our does not involve a mechanism to capture the surface-level information of an entity mention. We discuss this issue further in our error analysis (Section \ref{sec:errorAnalysis}).

\subsection{Ablation and Parameter Studies}

\subsubsection{Study of Sensitivity to Parameters}
\label{sec:paramSens}
\begin{figure}[ht]
    \centering
    \includegraphics[width=0.7\columnwidth]{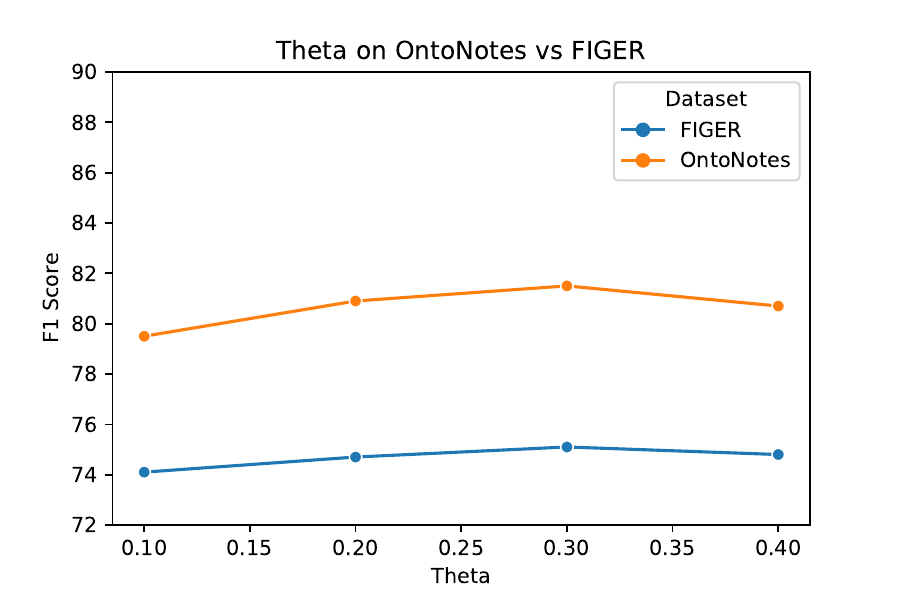}
    \vspace{-2ex}
    \caption{Parameter study: $\theta$ on OntoNotes and FIGER}
    \label{fig:Param}
    \vspace{-0.5cm}
\end{figure}

A potential concern with the experimental setup
can be overtly high sensitivity of \our to the Entity Type Specificity parameter $\theta$. For all experiments in Table \ref{tab:results}, we leverage the same $\theta$ value of 0.3. Additionally, from the plot in Figure \ref{fig:Param}, we clearly see that F1 scores are not drastically sensitive to theta with standard deviations of 0.3631 and 0.6262 on FIGER and OntoNotes respectively.

\subsubsection{\our Module Ablation Study}
\label{sec:abla}

\begin{figure}[ht]
\centering
\includegraphics[width=0.7\columnwidth]{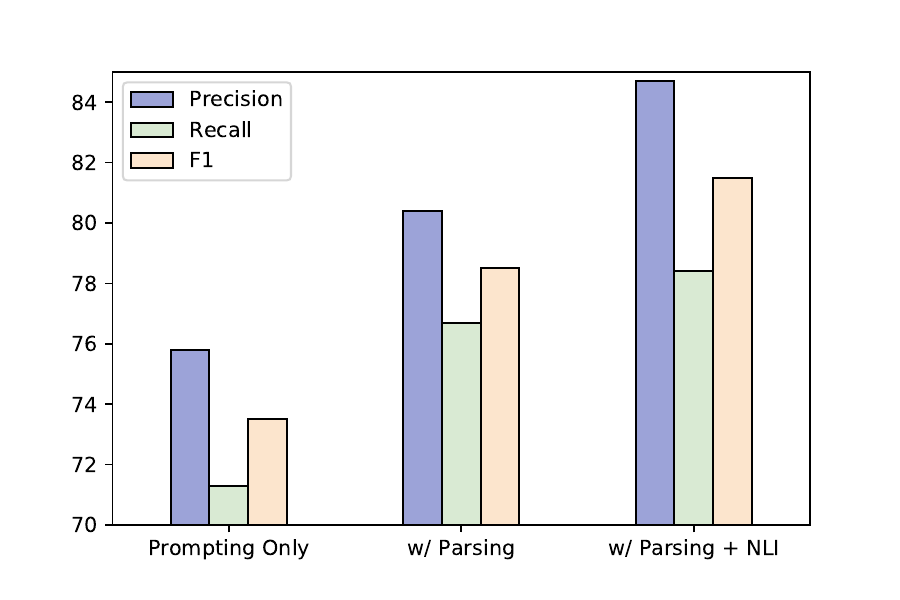}
    \caption{\our ablation study on OntoNotes: Results (\%)}
    \label{fig:Ablation}
\vspace{-0.5cm}
\end{figure}

We also include the prediction results of our ablation models to demonstrate how the NLI module contributes to the final type assignment. We utilize a simple type mapping to evaluate our ensembled BERT Cloze Prompting module. Figure \ref{fig:Ablation} shows the results of ablation studies on the test set of the Ontonotes dataset. We compared the \our full model with various ablations and variations. We find that direct prompting performs the worst whereas including parsed head words further improves the recall. Including the NLI module significantly improves the precision of the framework as we are able to leverage the type hierarchy to consolidate the fine-grained type that best represents the initial candidates. 
Furthermore, we include Table \ref{tab:AblationTypeAlignment} to examine alternative embedding approaches for the high-level type assignment. When performing high-level type assignment, we aim to leverage the surface-level information of the candidate type set to select the high-level type that is most conceptually similar. As a result, we find the one with word2vec embeddings performs better than BERT\cite{devlin-etal-2019-bert} and RoBERTa\cite{roberta} which incorporate the semantic context of a word in the sentence. 

\subsubsection{Efficiency \& Scalability Study}
Finally, we conduct an efficiency and scalability study to gauge the impact of larger scale data sets on our method and include the results in Table \ref{tab:Scalability}.

\begin{table}[h]
\centering
\begin{tabular}{ll}
\hline
    \textbf{Number of Mentions} & \textbf{Time to Predict Types}  \\ \hline
    1000 &  1.75 min \\ 
    25000 & 57.1 min \\ 
    100000 &  4.3 hr \\ \hline
\end{tabular}
\caption{Results of the efficiency \& scalability study.}
\label{tab:Scalability}
\vspace{-1cm}
\end{table}

\begin{table}[ht]
\small
\centering
\begin{tabular}{llll}
\hline
    \textbf{Model} & \textbf{Prec} & \textbf{Rec} & \textbf{Ma-F1}  \\ \hline
    $\our_{BERT}$ & 82.3 & 77.1 & 79.6  \\ 
   $\our_{RoBERTa}$ & 81.9 & 76.9 & 79.4  \\
   $\our_{Word2Vec}$ & \textbf{84.7} & \textbf{78.4} & \textbf{81.5}  \\ \hline
\end{tabular}
\caption{High-level type alignment ablation study: Results (\%)}
\label{tab:AblationTypeAlignment}
\vspace{-1cm}
\end{table}

\subsection{Comparative Case Study}

Table \ref{tab:Case1} 
presents a sentence from a recent news article with tagged mentions and predicted entity types. We find that methods like MZET sometimes predict incompatible types due to incorrect or misleading surface information. For example, when typing ``The White House'', MZET and ZOE leverage the surface-level information from large KBs to identify the mention as a location. However, given the local context, the White House clearly refers to the U.S. government. When considering the mention ``US President Joe Biden'', our method utilizes the type information from the candidate type set (Official, Leader, Politician, Individual) to explore the ``Person'' branch within the ontology, and then it searches the ``Politician'' branch to select the best fine-grained context-aware type ``President'' for our given entity mention. Thus, with the assistance of the PLMs, we can incorporate contextual information to derive more context-aware type labels. 

\section{Discussion }

\subsection{Modified Fine-Grained Type Ontology}
The structure of the fine-grained type ontology is important to the performance of \our. The input ontology must be built using hypernym-type relationships where each parent type is a generalization of the child type. The provided FIGER ontology contains logical inconsistencies in how various types of relations are organized. 

For example, the FIGER ontology considers the parent of the ``Road'' type to be ``Transportation'' rather than ``Location''. This logical inconsistency leads to erroneous typing from our method. 

Table \ref{tab:results} includes our results on the test set of FIGER with and without a modified type ontology. 

The modification of an ontology is done by reorganizing the types to leverage hypernym-hyponym type relationships. 
That is, in the revised type ontology, each parent type is a generalization of its child(ren) type(s). The included FIGER type ontology considers the parent type of the ``Road'' type to be "Other\textbackslash Transportation'' and the parent type of ``Building'' to be ``Other''. While ``Other'' can be considered a valid generalization, it is extremely broad and the coarse-grained type of ``Location'' serves as a stronger parent for both fine-grained types. Thus, we modify the included ontology by reorganizing the fine-grained types under parent types that have stronger generalizations (e.g., Building \& Road under Location rather than Other). \our achieves 2.1 absolute Ma-F1 improvement given this modified type ontology. 
Thus, we find that \our can be further improved with correct type ontologies to leverage the inherent hierarchical relationships between coarse and fine-grained types.
We provide insights into specific reasons for the mistakes made by the \our framework. For our analysis, we collect some of the erroneous decisions in the Ontonotes and FIGER test sets. We highlight the Gold Type for each mention in blue.

\subsection{Error Analysis: \our}
\label{sec:errorAnalysis}
\our, though having high performance, still generates nontrivial errors.
We analyze the reasons behind the errors on the Ontonotes and FIGER test sets and categorize them into three types. Note the Gold Type for each mention is highlighted in blue (and \textit{Italian font}).

\smallskip
\noindent
\textbf{Error Type 1: Lack of Knowledge Base} 

\smallskip
\noindent
\textbf{Sentence E1:} He was a caseworker in \textbf{Minnesota} \textit{\correct{[\textbackslash Location\textbackslash Region]}} but left the job because he found himself perpetually sick from the environments in which he worked. 

\vspace{0.3ex}
\noindent
In E1, \our incorrectly types Minnesota as a Country rather than the gold type of Region. In context, we should consider State a more reasonable type to predict. With a simple KB-matching method, we would be able to capture the surface-level information to type it as a State/Province. \our relies on neither human annotations nor a knowledge base.
Obviously, introducing a knowledge base and the KB-matching mechanism will further improve its performance.

\smallskip
\noindent
\textbf{Error Type 2: Incomplete Fine-Grained Type Ontology} 

\smallskip
\noindent
\textbf{Sentence E2:} Valley Federal Savings \& Loan Association said Imperial Corp. of America withdrew from regulators its application to buy \textbf{five Valley Federal branches} \textit{\correct{[\textbackslash Location\textbackslash Structure]}}, leaving the transaction in limbo.

\vspace{0.3ex}
\noindent
In E2, \our generates \textbackslash Other. Even when \our is able to generate high-quality candidate labels, it can sometimes fail to align to the correct entity type due to an incomplete type set. In this example, our Candidate Type set generated by prompting consists of Asset, Property, Facility, Bank, and Branch. 
Clearly, the best fine-grained type should be Asset or Property (rather than the Gold Type: Location or Structure).  
Since the provided OntoNotes ontology does not include such fine-grained types, \our is unable to generate the correct answer.
Clearly, a refined ontology will further improve its performance.

\smallskip
\noindent
\textbf{Error Type 3: Incapability to Type Nested Entities} 

\smallskip
\noindent
\textbf{Sentence E3:} The 33-year-old \textbf{Billings} \textit{\correct{[\textbackslash Location\textbackslash City]}} native enlisted as a military veterinarian.

\vspace{0.3ex}
\noindent
In E3, \our identifies Billings as \textbackslash Person. 
This mistake can be caused by confusing
the type of the whole entity ``Billings native'' with the type of the nested entity ``Billings''. 
PLM is good at generating candidate types for the whole entity based on its contextual structure, whereas the knowledge about a nested entity like ``Billings'' (in front of head word ``native'') can be more easily derived from a knowledge-base or from some nested entity type patterns.
We will leave the issue of typing nested entities to future work.

\subsection{Error Analysis: LLM on Fine-grained Typing}
\label{sec:LLMAnalysis}
Recent developments of large language models (LLMs) (e.g., ChatGPT and GPT-4 \cite{openai2023gpt4}) lead to enhanced capability at generating high-quality responses to prompts based on the knowledge learned from their extensive training corpora. 
Clearly, LLMs can provide stronger \emph{context-aware} knowledge for \our's candidate type generation.
However, without an ontological structure as guidance, an LLM (e.g. ChaptGPT) may still generate many errors, leading to lower performance than our method, as shown in Table \ref{tab:results}. Here we conduct error analysis for types generated by ChatGPT through the following prompt: \emph{“Return the fine-grained entity types of the given entity mentioned in the sentences below. Be concise and ONLY utilize the types from this list of possible entity types.
[entity types] \{entity types\} [sentence] \{sentence\} [entity mention] \{entity mention\}”}.

\vspace{0.3ex}
\noindent
\textbf{LLM Error Type 1: Incomplete Entity Types}\\
\textbf{LLM-E1}: Given a sentence $S_2$: ``\emph{The ceremony will take place Feb. 16–20.}'' and a parsed entity mention span ``\emph{Feb. 16–20}'', we prompted ChatGPT to generate candidate types for the masked entity mention span, which generates: ``Later'', ``There'', ``Outside'' and ``Indoors''. While these tokens complete the sentence accurately, they cannot serve as precise hypernyms or fine-grained types for the entity mention span of interest. With Hearst patterns, we can elicit more accurate hypernyms like ``Time'', ``Date'' and ``Event''. However, we are still unable to derive the most accurate fine-grained label of ``Period'' or ``Interval''. In contrast, \our utilizes a top-down approach guided by a fine-grained type ontology to refine and finally select the most conclusive entity type. 

\smallskip
\noindent 
\textbf{LLM Error Type 2: Incorrect Entity Types}

\vspace{0.3ex}
\noindent
\textbf{LLM-E2}: Given a sentence $S_3$: ``\emph{It will be the first time the Falcon Heavy has conducted a launch for the U.S. military’s secretive X-37B spaceplane project}'' and a parsed entity mention span ``\emph{Falcon Heavy}'', we can again prompt ChatGPT to generate candidate types for the masked entity mention span. With our prompt, ChatGPT generates: ``Company'', ``Organization'', ``Team'', ``Government'', and ``Agency''. With Hearst patterns, we can extract more accurate candidates like ``Vehicle'', ``Carrier'' and ``Launcher''. Furthermore, ChatGPT is unable to derive the most accurate fine-grained label of ``Spacecraft'', ``Shuttle'' or ``Rocket''. While some generated tokens accurately type the mention span, ChatGPT still requires a structured mechanism to mitigate hallucinations and select the most fine-grained entity type. 

Overall, to achieve high quality fine-grained typing, we believe an LLM should be assisted with structured knowledge
(e.g., a fine-grained ontological structure), which is an interesting direction for future research.

\section{Conclusions}

We propose \our, which leverages the weak supervision setting of pre-trained language model prompting. We use head words and MLM cloze prompting for fine-grained candidate label generation. Then we automatically match the generated fine-grained types to our type ontology with an inference method to select the most appropriate fine-grained types under the local context. Extensive experiments on real-world datasets show that \our is highly effective and substantially outperforms the state-of-the-art zero-shot FET methods. In the future, we plan to further refine and enrich a type ontology which will enable us to incorporate more type information for even better performance. Furthermore, \our lacks the capability to address nested entities (e.g., \emph{Denver} Native). We plan to resolve this in future work by incorporating boundary knowledge to extract accurate and complete type information. 

\section*{Acknowledgement}
Research was supported in part by the Molecule Maker Lab Institute: An AI Research Institutes program supported by NSF under Award No. 2019897, US DARPA KAIROS Program No. FA8750-19-2-1004, INCAS Program No. HR001121C0165, National Science Foundation IIS-19-56151, and the Institute for Geospatial Understanding through an Integrative Discovery Environment (I-GUIDE) by NSF under Award No. 2118329. Any opinions, findings, and conclusions or recommendations expressed herein are those of the authors and do not necessarily represent the views, either expressed or implied, of National Science Foundation, DARPA or the U.S. Government. The views and conclusions contained in this paper are those of the authors and should not be interpreted as representing any funding agencies.
\newpage

\bibliographystyle{ACM-Reference-Format}
\vfill\eject
\bibliography{main}
\appendix
\section{Evaluation Metrics}
\label{sec:evalMetrics}
Following the prior FET studies (\cite{dai-etal-2021-ultra}, \cite{Weld_Ling2012}, \cite{ma-etal-2016-label}), we evaluate our methods and the baselines using three evaluation metrics: Strict Accuracy (Acc), Micro-F1 (Mi-F1), and Macro-F1 (Ma-F1). 

\noindent  
{\sf Accuracy.}
Given a set of entity mentions $M$, we denote the set of ground truths and predicted types as $t_M$ and $\hat{t_M}$ respectively. Given $\sigma$ as an indicator function, strict accuracy is defined as 
$$Acc = \frac{\Sigma_{m\in M} \sigma(t_m == \hat{t_m})}{M}$$ 
 
\noindent  
{\sf Macro-F1.}
Macro-F1 is calculated using Macro-Precision ($P_{ma}$) and Macro-Recall ($R_{ma}$) where
$$P_{ma} = \frac{1}{|M|} \Sigma_{m\in M} \frac{|t_m \cap \hat{t_m}|}{\hat{t_m}}$$
$$R_{ma} = \frac{1}{|M|} \Sigma_{m\in M} \frac{|t_m \cap \hat{t_m}|}{t_m}$$

\noindent  
{\sf Micro-F1.}
Micro-F1 is calculated using Micro-Precision ($P_{mi}$) and Micro-Recall ($R_{mi}$) where 
$$P_{mi} = \frac{\Sigma_{m\in M} |t_m \cap \hat{t_m}|}{\Sigma_{m\in M} \hat{t_m}}$$
$$R_{mi} =   \frac{\Sigma_{m\in M} |t_m \cap \hat{t_m}|}{\Sigma_{m\in M} t_m}$$
Macro-F1 and Micro-F1 are calculated using the F1 score formula with their respective granular precision and recall scores.
$$F_1 = \frac{2*Precision*Recall}{Precision+Recall}$$
\vfill\eject
\end{document}